\documentclass[a4paper, 10pt, conference]{ieeeconf} 

\IEEEoverridecommandlockouts    

\usepackage{graphics} 
\usepackage{epsfig} 
\usepackage{mathptmx} 
\usepackage{times} 
\usepackage{amsmath} 
\usepackage{amssymb} 
\usepackage{color}
\usepackage{xcolor}
\usepackage{booktabs,siunitx} 
\usepackage{svg}
\usepackage{multirow}
\usepackage{rotating}
\usepackage[export]{adjustbox}
\usepackage{array, makecell}
\usepackage{hyperref}

\usepackage[symbol]{footmisc}

\title{\LARGE \bf
Semantic Understanding of Foggy Scenes with Purely Synthetic Data 
}

\newcommand\Tstrut{\rule{0pt}{2.2ex}}         

\author{Martin Hahner$^{1}$, Dengxin Dai$^{1}$, Christos Sakaridis$^{1}$, Jan-Nico Zaech$^{1}$, and Luc Van Gool$^{1,2}$
\thanks{$^{1}$Martin Hahner, Dengxin Dai, Christos Sakaridis, Jan-Nico Zaech and Luc Van Gool are all with the Toyota TRACE-Zurich team at the Computer Vision Lab, ETH Zurich, 8092 Zurich, Switzerland
 {\tt\small firstname.lastname@vision.ee.ethz.ch}}%
\thanks{$^{2}$Luc Van Gool is also with the Toyota TRACE-Leuven team at the Dept. of Electrical Engineering ESAT, KU Leuven, 3001 Leuven, Belgium
 {\tt\small luc.vangool@kuleuven.be}}%
}

\begin{document}

\maketitle
\thispagestyle{empty}
\pagestyle{empty}

\begin{abstract}
This work addresses the problem of semantic scene understanding under foggy road conditions. Although marked progress has been made in semantic scene understanding over the recent years, it is mainly concentrated on clear weather outdoor scenes. Extending semantic segmentation methods to adverse weather conditions like fog is crucially important for outdoor applications such as self-driving cars. In this paper, we propose a novel method, which uses purely synthetic data to improve the performance on unseen real-world foggy scenes captured in the streets of Zurich and its surroundings. Our results highlight the potential and power of photo-realistic synthetic images for training and especially fine-tuning deep neural nets. Our contributions are threefold, 1) we created a purely synthetic, high-quality foggy dataset of 25,000 unique outdoor scenes, that we call \textit{Foggy Synscapes} and plan to release publicly 2) we show that with this data we outperform previous approaches on real-world foggy test data 3) we show that a combination of our data and previously used data can even further improve the performance on real-world foggy data.

\end{abstract}

\section{Introduction} 

The last years have seen tremendous progress in tasks relevant to autonomous driving~\cite{drive:surroundview:route:planner}. It has also been hyped that autonomous vehicles of multiple companies have driven for several millions of miles by now. This evaluation or measurement, however, is mainly performed under favorable weather conditions such as the typically great weather in California. In the meanwhile, the development of computer vision algorithms are also focused and benchmarked with clear weather images. As argued in \cite{vision:atmosphere,SFSU_synthetic},  outdoor applications such as automated cars, however, still need to function well in adverse weather conditions. One typical example of an adverse weather condition is fog, which degrades data quality and thus the performance of popular perception algorithms significantly. The challenge exists for both Cameras \cite{contrast:weather:degraded,SFSU_synthetic} and LiDAR sensors \cite{lidar:fog:itsc:16,Benchmarking_Adverse}. This work investigates semantic understanding of foggy scenes with Camera data. 

Currently, in the era of deep learning, the most popular and best performing algorithms addressing semantic scene understanding are neural networks trained with many annotations of real images~\cite{imagenet:2015,Cityscapes}. While this strategy seems to be promising as many algorithms still benefit from having more data, applying the same protocol to all adverse conditions (e.g. fog, rain, snow and nighttime) and their combinations (e.g. foggy night) is problematic. The manual annotation part is hardly scalable to so many domains. This cost of manual annotation is more pronounced for adverse weather conditions, where it is---due to the poor visibility---much harder to provide precise human annotations. This paper aims at improving semantic understanding of real foggy scenes without using additional human annotations of real foggy images.  

To overcome this problem, Sakaridis et al. \cite{SFSU_synthetic} has recently proposed an approach to imposing synthetic fog into real clear weather images and learning with those partially synthetic data. While it generates state of the art results for the task, the method has a few drawbacks. First, the size of the generated partially synthetic dataset is limited by the size of existing datasets created for clear weather condition as the ground truth labels are inherited from the latter. Furthermore, their method requires depth completion and de-noising of real-world scenes which itself is a very challenging and unsolved problem. The imperfect depth maps lead to artifacts in simulated fog. In order to address these two issues, this work takes a step further and develops a method for semantic foggy scene understanding with purely synthetic data. By purely synthetic data, we mean that both the underlying images and the imposed fog are synthetic. Due to the synthetic nature, the images, its corresponding semantic labels and its corresponding depth maps can be obtained easily via running rendering algorithms. More importantly, the depth maps are accurate, which leads to realistic fog simulation. While our method addresses the two problems of \cite{SFSU_synthetic}, the drawback lies in its synthetic nature of underlying scenes, which may lack the richness of real-world scenes.  

The main aim of this work is to answer the following two  questions: (1) whether purely synthetic data can outperform partially synthetic data for semantic understanding of real foggy scenes; and (2) whether these two complementary data sources (one with better underlying images and the other with better fog effect) can be combined and boost the performance further. The short answer to both questions is yes and the detailed answers are given in the following sections. 

To summarize, the main contributions of the paper are: 1) proposing a new purely synthetic dataset for semantic foggy scene understanding which features $25 000$ high-resolution foggy images; 2) demonstrating that purely synthetic data of foggy scenes with accurate depth information can outperform partially real data with imperfect depth information when tested on real foggy scenes; and 3) demonstrating that the combination of purely synthetic fog data and partially synthetic fog data gives the best results than either of the method alone. 

Our work takes advantage of the recent progress in computer graphics for generating realistic synthetic driving data \cite{Synthia:dataset,playing:data,Synscapes}. It also further reinforces the current belief that there is a great potential of learning with high-quality synthetic data. 

\textit{Foggy Synscapes} will be publicly available at \\ \url{trace.ethz.ch/foggy_synscapes}.

\section{Related Work}

A large body of recent literature is dedicated to semantic understanding of outdoor scenes under \textit{normal} weather conditions, developing both large-scale annotated datasets~\cite{kitti,Cityscapes,Mapillary} and end-to-end trainable models~\cite{dilated:convolution,refinenet,pspnet,DeepLabv3+} which leverage these annotations to create discriminative representations of the content of such scenes. However, most outdoor vision applications, including autonomous vehicles, also need to remain robust and effective under \textit{adverse} weather or illumination conditions, a typical example of which is fog~\cite{vision:atmosphere}. The presence of fog degrades the visibility of a scene significantly~\cite{contrast:weather:degraded,tan2008visibility} and the problem of semantic understanding becomes more severe as the fog density increases. Even though the need for specialized methods and datasets for semantic scene understanding under adverse conditions has been pointed out early on in the literature~\cite{Cityscapes}, only very recently has the research community responded to this need both in the dataset~\cite{wilddash,night:stylized:uncertainty} and in the methodological direction~\cite{SFSU_synthetic,dense:SFSU:eccv18,daytime:2:nighttime,incremental:adversarial:DA:18,Benchmarking_Adverse, SeeingThroughFog}. Our work also answers this need and specifically targets the condition of fog, by constructing a large-scale photo-realistic synthetic foggy dataset that originates from a synthetic clear weather counterpart to enable adaptation to fog.

The SYNTHIA~\cite{Synthia:dataset} and GTA~\cite{playing:data} datasets are the first examples of purely synthetic data---rendered with video game engines---that were used for training in combination with real data to improve semantic segmentation performance on real outdoor scenes, while similar work concurrently considered indoor scenes~\cite{indoor:synthetic}. The main advantage of these approaches is the drastically reduced cost of ground-truth generation compared to real-world datasets, which require demanding manual annotation. The utility of purely synthetic training data has been further emphasized in~\cite{driving:in:the:matrix} for the task of vehicle detection, where a model trained \textit{only} on a massive synthetic dataset is shown to outperform the corresponding model trained on the real large-scale Cityscapes~\cite{Cityscapes} dataset. While all aforementioned works on synthetic data pertain to \textit{normal} conditions, Sakaridis et al. \cite{SFSU_synthetic} generated Foggy Cityscapes, a partially synthetic foggy dataset created by simulating fog on the original clear weather scenes of Cityscapes~\cite{Cityscapes} and inheriting its ground-truth semantic annotations at no extra cost. The fog simulation pipeline of~\cite{SFSU_synthetic} is improved in~\cite{dense:SFSU:eccv18} by leveraging semantic annotations to increase the accuracy of the required depth map, resulting in the Foggy Cityscapes-DBF dataset. Both of these synthetic foggy datasets have been utilized in~\cite{SFSU_synthetic,dense:SFSU:eccv18} to improve semantic segmentation performance of state-of-the-art CNN models~\cite{dilated:convolution,refinenet} on \textit{real} foggy benchmarks. We are inspired by both lines of research and combine the fully controlled setting of purely synthetic data with the synthetic fog generation pipeline of~\cite{SFSU_synthetic,dense:SFSU:eccv18}. In particular, we exploit the very recent large-scale photo-realistic Synscapes~\cite{Synscapes} dataset, which comes with ground-truth depth maps, to generate Foggy Synscapes. The ground-truth depth in Synscapes drastically simplifies the fog simulation pipeline and completely eliminates artifacts in the resulting synthetic images of Foggy Synscapes. By contrast, potentially incorrect estimation of depth values in the complex pipeline of~\cite{SFSU_synthetic,dense:SFSU:eccv18} introduces artifacts to Foggy Cityscapes and Foggy Cityscapes-DBF.

Besides the above recent works on pixel-level parsing of foggy scenes, there have also been earlier works on fog detection~\cite{fog:detection:cv:09,fog:detection:vehicles:12,night:fog:detection,fast:fog:detection}, classification of scenes into foggy and fog-free~\cite{fog:nonfog:classification:13}, and visibility estimation both for daytime~\cite{visibility:road:fog:10,visibility:detection:fog:15,fog:detection:visibility:distance} and nighttime~\cite{night:visibility:analysis:15}, in the context of assisted and autonomous driving. The closest of these works to ours is~\cite{visibility:road:fog:10}, which generates synthetic fog and segments foggy images to free-space area and vertical objects. However, our semantic segmentation task lies on a higher level of complexity and we employ state-of-the-art CNN architectures, exploiting the most recent advances in this area.

Our work also bears resemblance to domain adaptation methods. \cite{road:scene:2013} focuses on adaptation across weather conditions to parses simple road scenes. More recently, adversarial domain adaptation approaches have been proposed for semantic segmentation, both at pixel level and feature level, adapting models from simulated to real environments~\cite{adversarial:training:simulated:17,synthetic:semantic:segmentation,CyCADA,incremental:adversarial:DA:18}. Our work is complementary, as we generate high-quality synthetic foggy data from photo-realistic synthetic clear weather data to enable adaptation to the foggy domain.

\section{Method}

\begin{figure*}
    \centering
    \hspace{0.5cm}
    \includegraphics[width=\textwidth]{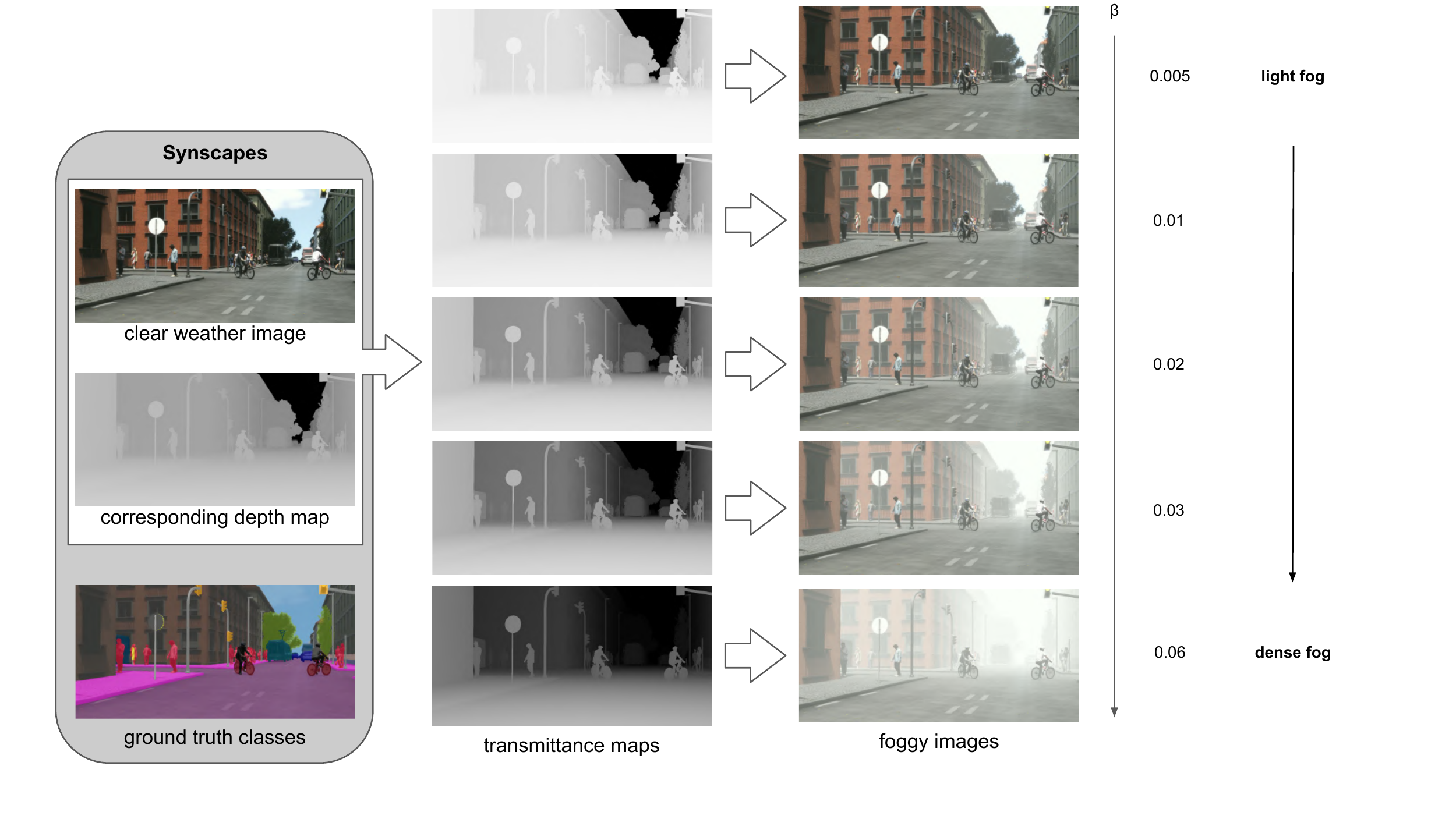}
    \caption{Our synthetic fog generation pipeline using the gapless depth information provided by the Synscapes~\cite{Synscapes} dataset.}
    \label{fig:simulation:pipeline}
\end{figure*}

In this section, we outline our synthetic fog generation pipeline displayed in Figure~\ref{fig:simulation:pipeline}. For more details, we refer the reader to~\cite{SFSU_synthetic}, where our pipeline is adapted from. 

The standard optical model for fog that forms the basis of this pipeline was introduced in~\cite{Koschmieder:optical:model} and is expressed as

\begin{equation} \label{eq:fog:model}
\mathbf{F}(\mathbf{x}) = t(\mathbf{x})\mathbf{R}(\mathbf{x}) + (1 - t(\mathbf{x}))\mathbf{L},
\end{equation}

where $\mathbf{F}(\mathbf{x})$ is the observed foggy image at pixel $\mathbf{x}$. $\mathbf{R}(\mathbf{x})$ is the clear scene radiance and $\mathbf{L}$ the atmospheric light, which is assumed to be globally constant. For the atmospheric light estimation we use the same approach as in~\cite{SFSU_synthetic}. For homogeneous fog, the transmittance depends on the distance $\ell(\mathbf{x})$ between the camera and the scene through

\begin{equation} \label{eq:transmittance}
t(\mathbf{x}) = \exp\left(-\beta\ell(\mathbf{x})\right).
\end{equation}

\begin{figure}[ht]
    \centering
    \includegraphics[width=0.55\columnwidth]{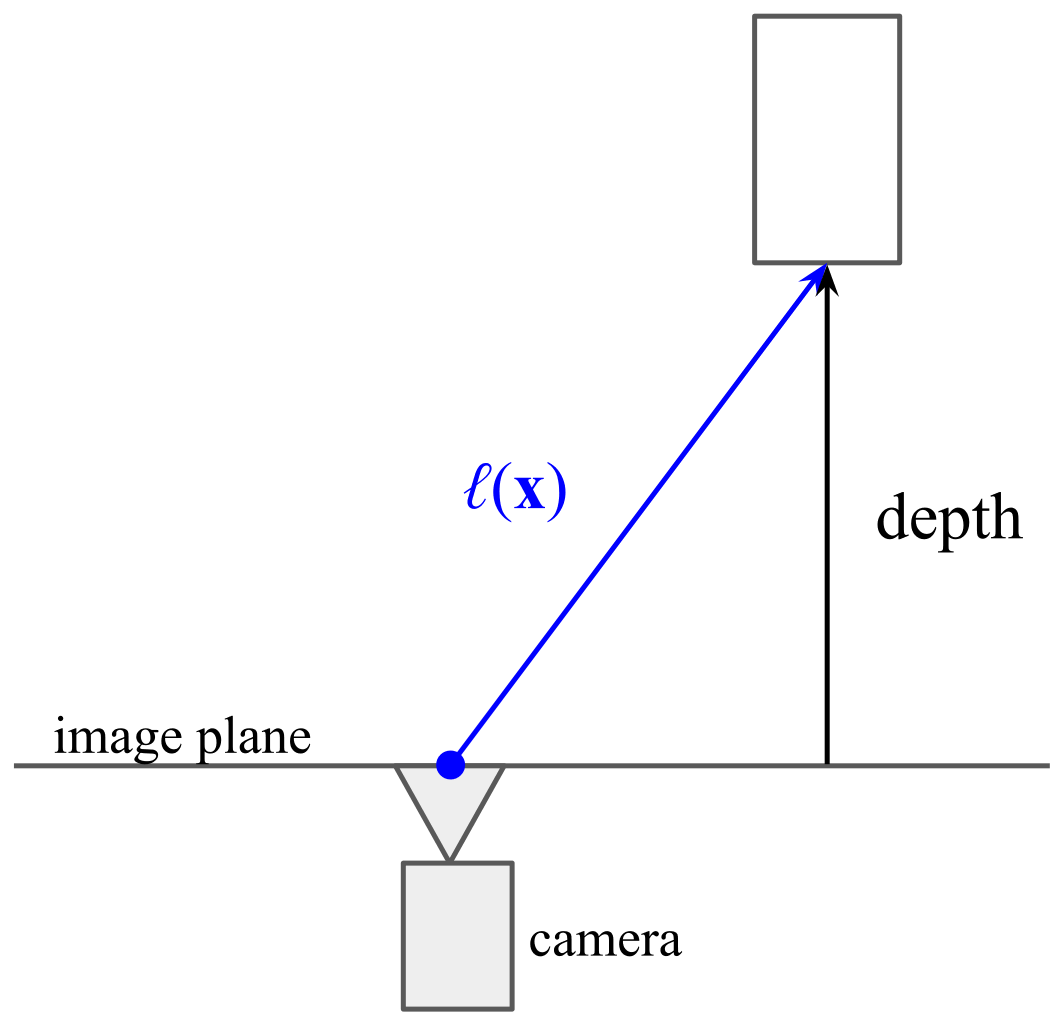}
    \caption{Difference between $\ell(\mathbf{x})$ and the commonly provided depth information in datasets like Cityscapes~\cite{Cityscapes} and Synscapes~\cite{Synscapes}.}
    \label{fig:illustration:camera}
\end{figure}

Note that the distance $\ell(\mathbf{x})$ in~(\ref{eq:transmittance}) is \textit{not} equivalent to the depth information provided in datasets that focus on automated driving like Cityscapes~\cite{Cityscapes} or Synscapes~\cite{Synscapes}. The provided depth in those datasets commonly measure the distance between the image plane and the scene. See Figure~\ref{fig:illustration:camera} for an illustration of the difference.

Further, the attenuation coefficient $\beta$ controls the density of the fog where larger values of $\beta$ correspond to denser fog. By definition of the National Oceanic and Atmospheric Administration within the U.S. Department of Commerce~\cite{Federal:meteorological:handbook} fog is called fog if it decreases the visibility, among meteorologist more formally called the \textit{meteorological optical range} (MOR), to less than $1\,\text{km}$. For homogeneous fog $\text{MOR}=\frac{2.996}{\beta}$ always holds.

So by the aforementioned definition ~\cite{Federal:meteorological:handbook}, the lower bound $\beta \geq 0.002996$ corresponds to the lightest fog configuration possible and as a matter of course is always obeyed in our synthetic fog generation pipeline, where $\beta~\in~[0.005, 0.01, 0.02, 0.03, 0.06]$ is used. These $\beta$-values correspond to a visibility of approximately $600\text{m}$, $300\text{m}$, $150\text{m}$, $100\text{m}$, and $50\text{m}$ respectively. 

In contrast to prior work ~\cite{SFSU_synthetic,dense:SFSU:eccv18} only using real input images (with incomplete and imperfect depth information) to the synthetic fog generation pipeline, our work focuses on synthetic input images (with complete and perfect depth information). Hence, our resulting images are purely synthetic.

For our experiments, we chose the recently released Synscapes~\cite{Synscapes} dataset. Synscapes is created by an end-to-end approach focusing on photo-realism using the same physically based rendering techniques that power high-end visual effects in the film industry. Those rendering techniques accurately capture the effects of everything from illumination by sun and sky, to the scene's geometry and material composition, to the optics, sensor and processing of the camera system. The dataset consists of $25 000$ procedural and unique clear weather images that do not follow any path through a given virtual world. Instead, an entirely unique scene is generated for each and every individual image. As a result, the dataset contains a wide range of variations and unique combinations of features.

\begin{figure*}
    \centering
    \includegraphics[width=\textwidth]{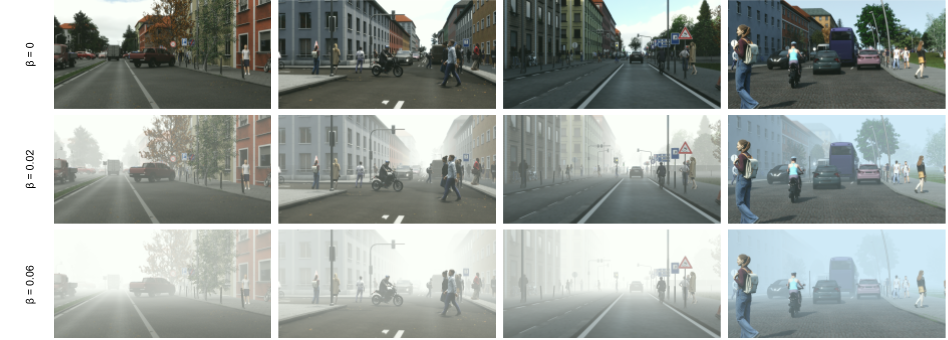}
    \caption{Comparison of clear weather images from Synscapes~\cite{Synscapes} against images from our adapted Foggy Synscapes for $\beta = 0$, $0.02$, and $0.06$.}
    \label{fig:fog:simulation}
\end{figure*}

Results of the presented pipeline for synthetic fog generation on example images from Synscapes~\cite{Synscapes} are provided in Figure~\ref{fig:fog:simulation} for $\beta = 0$ (which is equivalent to the clear weather input image), $0.02$ and $0.06$. $\beta$-values of $0.02$ and $0.06$ corresponds to a visibility of approximately $150\text{m}$ and $50\text{m}$ respectively. The required inputs in~(\ref{eq:fog:model}) are the clear weather image $\mathbf{R}$, the atmospheric light $\mathbf{L}$ and the corresponding transmittance map $t$. We call this new dataset Foggy Synscapes, where the ground truth annotations are inherited as is from Synscapes.

In the rightmost column in Figure~\ref{fig:fog:simulation}, where there are no clouds present in the sky of the clear weather image, we can also see a rare failure case of our synthetic fog generation pipeline. Due to the missing clouds, in this image, a pixel of the blue sky will be selected as atmospheric light constant, which leads to the bluish tint in the synthetic fog. This is where our assumption of the air being totally homogeneous breaks. In images like the one in the second-rightmost column in Figure~\ref{fig:fog:simulation}, when there is blue sky and clouds, our pipeline does not break since it will pick the atmospheric light constant from a pixel in the clouds. 

\begin{figure*}[tb]
    \centering
    \includegraphics[width=.9\textwidth]{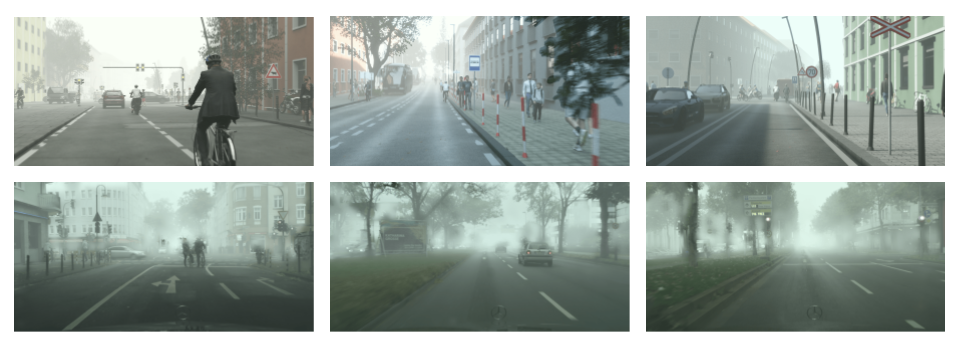}
    \caption{Qualitative comparison between our Foggy Synscapes (top) and Foggy Cityscapes~\cite{SFSU_synthetic} (bottom) for $\beta = 0.02$.}
    \label{fig:fog:comparison}
\end{figure*}

In Figure~\ref{fig:fog:comparison}, we qualitatively compare our Foggy Synscapes to Foggy Cityscapes~\cite{SFSU_synthetic}. One can see that the synthetically added fog in our Foggy Synscapes looks far more realistic than the synthetically added fog in Foggy Cityscapes. To a great extent, this is due to the perfect depth information provided by the original Synscapes~\cite{Synscapes} dataset. If the provided depth was not accurate, the quality of the synthesized foggy images would degrade and we would have artifacts similar to the ones present in Foggy Cityscapes.

\section{Experiments}

In this section, we present our findings on two real-world datasets that contain foggy scenes of various densities. The first one is called Foggy Driving~\cite{SFSU_synthetic}. A dataset which is exclusively meant for testing. It contains $101$ annotated images for all $19$ evaluation classes of Cityscapes~\cite{Cityscapes}. While $33$ images are finely annotated for every pixel in the image, the majority of images ($68$ images) are annotated at a coarser level. $51$ of the images were captured with a cell phone camera in foggy conditions at various areas of Zurich and the remaining $50$ images were collected from the web. Foggy Driving~\cite{SFSU_synthetic} contains more than $500$ annotated vehicles and almost $300$ annotated humans.

The second dataset is called Foggy Zurich~\cite{dense:SFSU:eccv18}, containing $3808$ real-world foggy road scenes captured while driving in the city of Zurich and its suburbs using a GoPro~Hero~5 attached to the inside of a car's windshield. Initially, its test split Foggy Zurich-test consisted of $16$ images with pixel-level semantic annotations for $18$ out of $19$ evaluation classes of Cityscapes~\cite{Cityscapes} (the train class is missing). In a more recent work~\cite{CMAda}, Foggy Zurich-test has been extended and now includes pixel-level semantic annotations for in total $40$ images. These $40$ images form the test set that we used for our evaluation. Compared to Foggy Driving, Foggy Zurich-test only includes foggy images of uniform and high resolution that are all annotated at a \textit{fine} level.

On the network architecture side, we chose RefineNet~\cite{refinenet} to compare with previous work~\cite{SFSU_synthetic} and confirmed our findings with a state of the art real-time semantic segmentation architecture BiSeNet~\cite{BiSeNet}. Regarding RefineNet, we used the same training and fine-tuning policy as in~\cite{SFSU_synthetic}. As second architecture we chose BiSeNet~\cite{BiSeNet} because we believe that lightweight real-time network architectures like BiSeNet~\cite{BiSeNet}, which have an order of magnitude less parameters than RefineNet~\cite{refinenet}, could be the way to tackle various adverse weather conditions such as haze and fog in the future. 

The baseline model of the BiSeNet~\cite{BiSeNet} architecture was trained for $80$ epochs on $2,975$ clear weather images from the Cityscapes~\cite{Cityscapes} training set using stochastic gradient descent, an initial learning rate of $0.01$ and polynomial learning rate decay with power $0.9$ as described in~\cite{pspnet}. Further, we used momentum $0.09$ and weight decay $0.0005$. The training always was carried out with a batch size of $4$ on a single NVIDIA~Titan~Xp. Fine-tuning with foggy images produced by our pipeline for all experiments described below was carried out with a $10\times$ lower initial learning rate ($0.001$) and the remaining hyper-parameters unchanged.

\subsection{Partially vs. Purely Synthetic Data}

For this experiment we fine-tuned the baseline models of RefineNet~\cite{refinenet} and BiSeNet~\cite{BiSeNet} once with the partially synthetic Foggy Cityscapes~\cite{SFSU_synthetic} and once with our purely synthetic Foggy Synscapes. For fine-tuning on Foggy Cityscapes, we chose the \textit{refined} subset of $498$ images, which are of better quality than the complete set of $2,975$ training images. Fine-tuning on this \textit{refined} set of Foggy Cityscapes was carried out for $50$ epochs and validation was executed every epoch ($498$ images). Only the model with the lowest validation loss was saved for testing on real-world foggy data. Fine-tuning on our Foggy Synscapes was conducted for one epoch of $24 500$ training images and $500$ images were excluded from training and kept as validation set. Validation using our dataset was executed every $125$ iterations ($500$ images) and only the model snapshot with the lowest validation loss was saved for testing here, too. 

In both, Table~\ref{table:foggy_driving_results} for Foggy Driving and in Table~\ref{table:foggy_zurich_results} for Foggy Zurich-test, we see that fine-tuning on purely synthetic data outperforms fine-tuning on partially synthetic data for both network architectures. 

\subsection{Quantity vs. Quality}

To even get better numbers on Foggy Synscapes, one could try to fine-tune not only for one, but maybe more epochs. For this paper, however, we wanted to go another way and wished to answer the question whether the benefit of our Foggy Synscapes lies only in its much larger quantity of images or whether it actually lies within its superior fog quality. Therefore we fine-tuned the baseline models of both network architectures only on the first $498$ images of Foggy Synscapes. Results on both datasets, presented in Table~\ref{table:foggy_driving_results} for Foggy Driving and Table~\ref{table:foggy_zurich_results} for Foggy Zurich-test, illustrate that the benefit truly lies within the quality of the synthetic fog and not just in the much larger scale of the dataset. 

Why some experiments with less images are even outperforming the experiment on the full size of Synscapes is a bit surprising. It could be that the first $498$ images we selected contain less error cases as the one visualized in the rightmost column of Figure~\ref{fig:fog:simulation}. This investigation of why exactly this is happening is left for future work. One could also imagine to explicitly filter out those failure cases by using the provided meta-data parameter \textit{sky\_contrast} of the original Synscapes~\cite{Synscapes} dataset. This parameter defines the contrast of the sky, where values between 2-3 indicate fully overcast sky and higher values between 5-6 indicate direct sunlight (which notably increase the chance of such failure cases).

\subsection{Combination of Partially \& Purely Synthetic Data}

Finally, we also investigate what happens if we fine-tune on the combination of both datasets, Foggy Cityscapes~\cite{SFSU_synthetic} and our Foggy Synscapes. Therefore we mixed the two datasets with a 2\,:\,1 ratio favouring our Foggy Synscapes, meaning for every two images of Foggy Synscapes, there is one Foggy Cityscapes image in the training and validation set. Ratios of 1\,:\,1 and 1\,:\,5 were also tested, but were not as beneficial as the ratio 2\,:\,1. Note that in this setting all images from Foggy Cityscapes had to be used multiple times since its \textit{refined} set of $498$ images is significantly smaller than our Foggy Synscapes with $24 500$ training images. Results of this experiment can also be seen in Table~\ref{table:foggy_driving_results} for Foggy Driving and Table~\ref{table:foggy_zurich_results} for Foggy Zurich-test. Improved performance on both datasets generally indicate that the mixture of both datasets is in parts significantly more powerful than one on its own.

\begin{table*}

\hfill
\resizebox{.945\textwidth}{!}{\begin{tabular}{rrlrrrrrrrrrrrrrrrrrrrr}
\rotatebox[origin=l]{90}{\thead{Model}}
&  \rotatebox[origin=l]{90}{Training}
&  \rotatebox[origin=l]{90}{\shortstack[l]{\hspace{-5pt}Fine-Tuning \\ \hspace{-5pt}(\# of images)}}
&  \rotatebox[origin=l]{90}{\thead{road}}   
&  \rotatebox[origin=l]{90}{\thead{sidewalk}} 
&  \rotatebox[origin=l]{90}{\thead{building}} 
&  \rotatebox[origin=l]{90}{\thead{wall}}   
&  \rotatebox[origin=l]{90}{\thead{fence}}   
&  \rotatebox[origin=l]{90}{\thead{pole}}   
&  \rotatebox[origin=l]{90}{\thead{tr.light}} 
&  \rotatebox[origin=l]{90}{\thead{tr.sign}} 
&  \rotatebox[origin=l]{90}{\thead{vegetation}} 
&  \rotatebox[origin=l]{90}{\thead{terrain}} 
&  \rotatebox[origin=l]{90}{\thead{sky}}   
&  \rotatebox[origin=l]{90}{\thead{person}} 
&  \rotatebox[origin=l]{90}{\thead{rider}}  
&  \rotatebox[origin=l]{90}{\thead{car}}    
&  \rotatebox[origin=l]{90}{\thead{truck}}  
&  \rotatebox[origin=l]{90}{\thead{bus}}    
&  \rotatebox[origin=l]{90}{\thead{train}}    
&  \rotatebox[origin=l]{90}{\thead{motorcycle}}
&  \rotatebox[origin=l]{90}{\thead{bicycle}} 
&  \rotatebox[origin=l]{90}{\thead{\textbf{mean IoU}}} \\

\hline 

\hline

\multirow{5}{*}{\rotatebox[origin=c]{90}{\shortstack{RefineNet \\ \cite{refinenet}}}} 

\Tstrut & C & -
& 90.1 & 29.3 & 68.3 & 27.3 & \textbf{16.7} & 41.3 & 54.2 & 59.6 & 68.0 &  6.8 & 88.7 & \textbf{60.9} & 45.4 & 66.4 &  5.5 &  9.6 & 45.4 &  9.8 & 48.4 & 44.3 \\ 

& C & FC \hspace{19pt}(498) 
& 91.7 & 29.7 & 73.0 & \textbf{29.0} & 14.8 & 43.4 & 54.0 & \textbf{61.6} & 71.2 &  6.9 & 85.7 & 59.3 & 46.7 & 67.3 &  8.4 & 17.2 & 53.7 & 13.1 & 48.9 & 46.1 \\

& C & FS \hspace{10pt}(24,500)

& \textbf{92.4} & 32.9 & \textbf{76.1} & 16.8 & 14.6 & 43.3 & 55.0 & 60.8 & 74.0 &  9.3 & 90.8 & 49.8 & 36.0 & \textbf{72.2} & 17.5 & \textbf{51.3 }& 65.0 & 11.1 & 50.3 & 48.4  \\

& C & FS \hspace{20pt}(498)
& 91.1 & \textbf{34.9}  & 74.4 & 18.4 & 15.7 & 43.7 & \textbf{57.0} & 60.0 & 75.8 & 10.0 & \textbf{92.7} & 55.5 & \textbf{47.3} & 71.8 & \textbf{24.8} & 30.0 & 48.0 & \textbf{17.3} & \textbf{54.4} & 48.6 \\

& C & FC + FS (498)
& \textbf{92.4} & 34.0 & \textbf{76.1} & 23.9 & 16.2 & \textbf{45.6} & 55.9 & \textbf{61.6} & \textbf{76.4} & \textbf{11.1} & 92.2 & 57.5 & 45.6 & 69.9 & 13.7 & 42.3 &\textbf{ 82.2} & 14.1 & 52.6 & \textbf{50.7} \\

\hline

\multirow{5}{*}{\rotatebox[origin=c]{90}{\shortstack{BiSeNet \\ \cite{BiSeNet}}}} 

\Tstrut & C & - 
& 85.1 & 21.5 & 46.9 & 6.2 & \textbf{13.1} & 12.1 & 24.4 & 31.9 & 61.0 & 1.8 & 66.2 & \textbf{43.6} & \textbf{17.1} & 39.3 & 0.3 & 12.7 & 1.4 & 0.0 & 32.2 & 27.2 \\ 

& C & FC \hspace{19pt}(498)  
& \textbf{88.0} & 23.7 & 56.0 & \textbf{23.8} & 7.4 & 16.2 & 31.9 & 32.7 & \textbf{68.3} & 0.8 & 79.1 & 42.2 & 16.4 & 50.8 & 0.2 & 13.7 & 5.8 & 0.0 & 18.4 & 30.3 \\ 

& C & FS \hspace{10pt}(24,500)
& 81.4 & 16.6 & 60.8 & 5.3 & 8.7 & 27.4 & 33.8 & 43.7 & 60.1 & 2.8 & 91.9 & 31.0 & 5.5 & 57.2 & 11.4 & 22.5 & 8.0 & 0.0 & 18.6 & 30.9 \\ 

& C & FS \hspace{20pt}(498)
& 82.0 & 18.4 & 67.9 & 6.0 & 8.9 & 26.6 & 37.4 & 41.7 & 64.7 & 1.2 & \textbf{93.3} & 30.1 & 2.0 & 52.1 & \textbf{11.7} & 18.4 & 14.7 & 0.0 & 27.4 & 31.8 \\ 

& C & FC + FS (498)
& 84.3 & \textbf{23.8} & \textbf{68.0} & 4.0 & 7.3 & \textbf{29.6} & \textbf{39.4} & \textbf{45.7} & 66.4 & \textbf{3.7} & 89.7 & 36.1 & 6.0 & \textbf{62.7} & 10.0 & \textbf{37.7} & \textbf{18.7} & 0.0 & \textbf{35.5} & \textbf{35.2} \\ 

\hline

\hline

\end{tabular}}
\caption{Test results on Foggy Driving. \protect\linebreak C = Cityscapes, FC = Foggy Cityscapes, FS = Foggy Synscapes}
\label{table:foggy_driving_results}

\hfill
\resizebox{.945\textwidth}{!}{\begin{tabular}{rrlrrrrrrrrrrrrrrrrrrrr}
\rotatebox[origin=l]{90}{\thead{Model}}
&  \rotatebox[origin=l]{90}{Training}
&  \rotatebox[origin=l]{90}{\shortstack[l]{\hspace{-5pt}Fine-Tuning \\ \hspace{-5pt}(\# of images)}}
&  \rotatebox[origin=l]{90}{\thead{road}}   
&  \rotatebox[origin=l]{90}{\thead{sidewalk}} 
&  \rotatebox[origin=l]{90}{\thead{building}} 
&  \rotatebox[origin=l]{90}{\thead{wall}}   
&  \rotatebox[origin=l]{90}{\thead{fence}}   
&  \rotatebox[origin=l]{90}{\thead{pole}}   
&  \rotatebox[origin=l]{90}{\thead{tr.light}} 
&  \rotatebox[origin=l]{90}{\thead{tr.sign}} 
&  \rotatebox[origin=l]{90}{\thead{vegetation}} 
&  \rotatebox[origin=l]{90}{\thead{terrain}} 
&  \rotatebox[origin=l]{90}{\thead{sky}}   
&  \rotatebox[origin=l]{90}{\thead{person}} 
&  \rotatebox[origin=l]{90}{\thead{rider}}  
&  \rotatebox[origin=l]{90}{\thead{car}}    
&  \rotatebox[origin=l]{90}{\thead{truck}}  
&  \rotatebox[origin=l]{90}{\thead{bus}}    
&  \rotatebox[origin=l]{90}{\thead{train}}    
&  \rotatebox[origin=l]{90}{\thead{motorcycle}}
&  \rotatebox[origin=l]{90}{\thead{bicycle}} 
&  \rotatebox[origin=l]{90}{\thead{\textbf{mean IoU}}} \\

\hline 

\hline

\multirow{5}{*}{\rotatebox[origin=c]{90}{\shortstack{RefineNet \\ \cite{refinenet}}}} 

\Tstrut & C & -
& 74.3 & 56.5 & 35.5 & 20.2 & 23.8 & 39.6 & 54.4 & 58.3 & 58.3 & 28.9 & 66.8 &  1.6 & \textbf{27.4} & 81.7 &  0.0 &  1.9 & \multicolumn{1}{c}{-} & 21.1 &  \textbf{6.2} & 34.6 \\ 

& C & FC \hspace{19pt}(498) 
& 81.2 & 56.7 & 36.5 & 27.5 & 24.6 & 44.2 & 59.6 & 57.8 & 48.2 & 33.6 & 50.2 &  5.3 & 25.3 & 81.9 &  0.0 & \textbf{29.2} & \multicolumn{1}{c}{-} & 36.0 &  3.1 & 36.9 \\ 

& C & FS \hspace{10pt}(24,500)
& 83.6 & 60.0 & 46.6 & 31.9 & 33.6 & 45.1 & 62.2 & 61.5 & 68.3 & 35.2 & 79.0 &  4.3 & 21.5 & \textbf{82.0} &  0.0 &  0.2 & \multicolumn{1}{c}{-} & 44.7 &  5.1 & 40.3
 \\

& C & FS \hspace{20pt}(498)
& \textbf{87.9} & 59.5 & \textbf{54.5} & 40.9 & \textbf{44.8} & 47.8 & \textbf{63.6} & \textbf{62.5} & \textbf{74.1} & \textbf{39.3} & \textbf{84.9} &  4.5 & 24.6 & 75.3 &  0.0 &  0.1 & \multicolumn{1}{c}{-} & 43.5 &  4.7 & \textbf{42.7} \\

& C & FC + FS (498)
& 87.5 & \textbf{60.6} & 46.0 & \textbf{41.1} & 38.5 & \textbf{48.2} & 62.4 & 61.9 & 67.3 & 38.1 & 74.4 &  \textbf{6.2} & 22.5 & 80.8 &  0.0 &  1.7 & \multicolumn{1}{c}{-} & \textbf{45.9} &  3.8 & 41.4  \\

\hline

\multirow{5}{*}{\rotatebox[origin=c]{90}{\shortstack{BiSeNet \\ \cite{BiSeNet}}}} 

\Tstrut & C & -
& 67.1 & 32.3 & 25.3 & 9.6 & 19.4 & 6.7 & 7.7 & 16.1 & 49.6 & 19.8 & 43.1 & 0.0 & 0.0 & 9.1 & 0.0 & 0.0 & \multicolumn{1}{c}{-} & 0.0 & 0.0 & 16.1 \\ 

& C & FC \hspace{19pt}(498) 
& 72.8 & 35.1 & 38.6 & 11.1 & 23.2 & 13.0 & \textbf{34.4} & 28.9 & 59.8 & 33.0 & 66.4 & 0.0 & \textbf{15.9} & 21.5 & 0.0 & 0.0 & \multicolumn{1}{c}{-} & \textbf{21.3} & 0.0 & 25.0 \\ 

& C & FS \hspace{10pt}(24,500)
& 71.6 & 36.6 & 52.4 & 28.8 & 25.6 & 17.4 & 26.2 & 38.0 & 65.6 & 38.8 & 87.7 & \textbf{0.7} & 1.7 & 37.0 & 0.0 & 0.0 & \multicolumn{1}{c}{-} & 0.0 & 0.0 & 27.8 \\ 

& C & FS \hspace{20pt}(498)
& 68.0 & 35.5 & \textbf{62.5} & 32.1 & 17.6 & 18.0 & 30.4 & 37.4 & 67.9 & 43.2 & \textbf{90.5} & 0.1 & 0.3 & 21.1 & 0.0 & 0.0 & \multicolumn{1}{c}{-} & 0.0 & 0.0 & 27.6 \\ 

& C & FC + FS (498)
& \textbf{81.1} & \textbf{41.5} & 60.3 & \textbf{33.5} & \textbf{28.5} & \textbf{21.8} & \textbf{34.4} & \textbf{40.5} & \textbf{68.0} & \textbf{48.2} & 87.9 & 0.2 & 1.1 & \textbf{39.0} & 0.0 & \textbf{0.1} & \multicolumn{1}{c}{-} & 0.0 & 0.0 & \textbf{30.9} \\ 

\hline

\hline

\end{tabular}}
\caption{Test results on Foggy Zurich-test. \protect\linebreak C = Cityscapes, FC = Foggy Cityscapes, FS = Foggy Synscapes}
\label{table:foggy_zurich_results}
\end{table*}

\subsection{Discussion}
All results presented are consistently achieved using $\beta~=0.01$~for Foggy Cityscapes~\cite{SFSU_synthetic} and $\beta = 0.005$ for our Foggy Synscapes. For the experiment where we combined both datasets, best results are achieved with $\beta = 0.005$. Higher $\beta$-values were also explored, but its results were not as beneficial as the $\beta$-values $0.005$ and $0.01$. Experiments using $\beta~\in~[0.03, 0.06]$ sometimes even showed a drop in performance compared to the clear weather baseline model. 

\begin{figure*}[tb]

    \includegraphics[width=.94\textwidth, right]{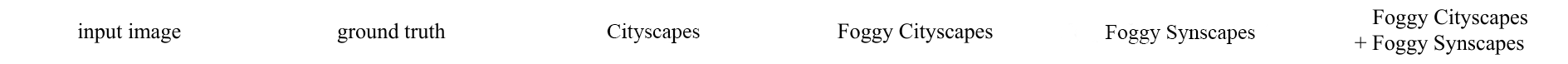}

    \begin{minipage}[c]{0.05\textwidth}
        \begin{turn}{90}
            RefineNet~\cite{refinenet}
        \end{turn}
    \end{minipage}
    \begin{minipage}[c]{0.94\textwidth}
        \includegraphics[width=\textwidth]{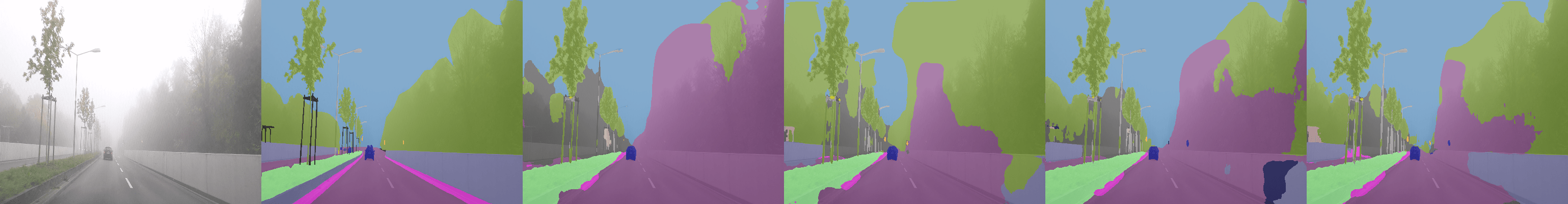} \\
        
        \includegraphics[width=\textwidth]{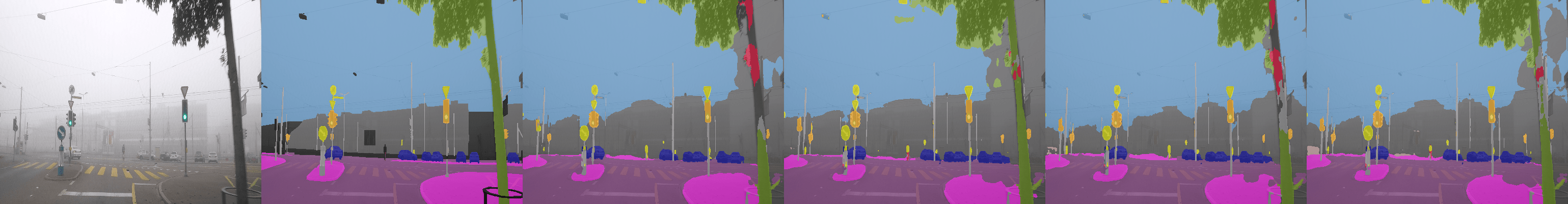} \\
    \end{minipage}
    
    \begin{minipage}[c]{0.05\textwidth}
        \begin{turn}{90}
            BiSeNet~\cite{BiSeNet}
        \end{turn}
    \end{minipage}
    \begin{minipage}[c]{0.94\textwidth}
        \includegraphics[width=\textwidth]{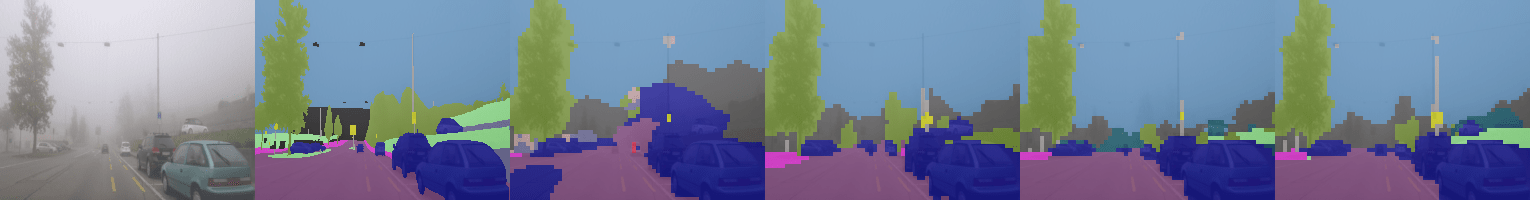} \\
        
        \includegraphics[width=\textwidth]{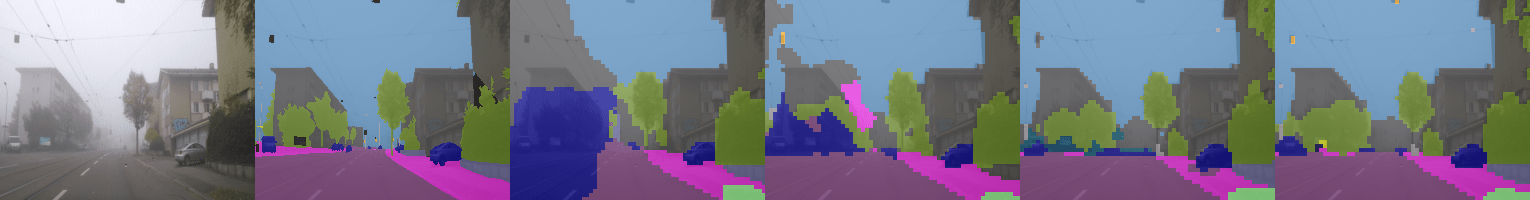} \\
    \end{minipage}
    
    \includegraphics[width=.947\textwidth, right]{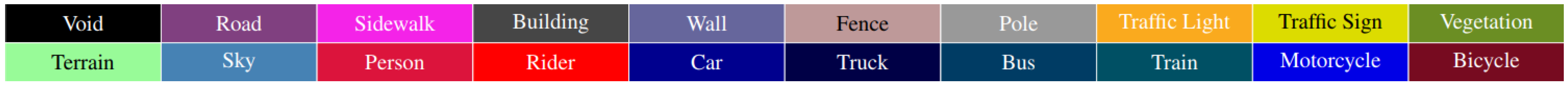}
  
    \caption{Representative RefineNet~\cite{refinenet} predictions (top two rows) and BiSeNet~\cite{BiSeNet} predictions (bottom two rows) on input images from Foggy Zurich predicted by the Cityscapes baseline model, alongside the predictions of the same model fine-tuned on Foggy Cityscapes, Foggy Synscapes, and the combination of both. This figure is best viewed in color and on screen.}
    \label{fig:foggy_zurich_results}
\end{figure*}

Figure~\ref{fig:foggy_zurich_results} shows four exemplary images of Foggy Zurich-test that show representative predictions using our Foggy Synscapes dataset. The top two rows illustrate results using the RefineNet~\cite{refinenet} and the bottom two rows illustrate results using the BiSeNet~\cite{BiSeNet} architecture. While improvements can be observed, the performance on foggy scenes is still a lot worse compared to what other papers may illustrate on clear weather scenes. This supports our claim that foggy scenes are indeed (way) more challenging than clear weather scenes.

Since our Foggy Synscapes is so much larger in scale than Foggy Cityscapes~\cite{SFSU_synthetic}, we also tried to train models directly on Foggy Synscapes (starting from ImageNet~\cite{imagenet:2015} pretrained weights). But unfortunately the resulting models could not outperform the clear weather \textit{Cityscapes} baseline models on real-world foggy data. So we concluded that the impact of real-world texture as present in Cityscapes~\cite{Cityscapes} seems to be still more important for the model "core" than the appearance of photo-realistic fog. 

\section{Conclusion}

In this paper, we were able to demonstrate that purely synthetic data of foggy scenes with accurate depth information can outperform partially real data with imperfect depth information when tested on real foggy scenes. We also showed that a combination of both types of data, purely synthetic and partially real, can further improve the scene understanding of real world foggy scenes. This is because in the setting where we merge the two datasets, the best of both worlds is combined. On the one side we have Foggy Cityscapes~\cite{SFSU_synthetic}, which contains real textures from Cityscapes~\cite{Cityscapes} but imperfect fog and on the other side we have our Foggy Synscapes, which has imperfect texture but much better looking fog than Foggy Cityscapes. We also learned that the power of Foggy Synscapes does not lie in its larger quantity, but mainly in its better fog quality. This further supports the assumption by the research community that there is great potential of evermore realistic looking synthetic data. 

\bibliographystyle{IEEEtran}

\end{document}